# Super-resolution using Sparse Representations over Learned Dictionaries: Reconstruction of Brain Structure using Electron Microscopy

Tao Hu, Juan Nunez-Iglesias, Shiv Vitaladevuni, Lou Scheffer, Shan Xu, Mehdi Bolorizadeh,
Harald Hess, Richard Fetter, Dmitri Chklovskii

Janelia Farm Research Campus, Howard Hughes Medical Institute

## Abstract

*A central problem in neuroscience is reconstructing neuronal circuits on the synapse level. Due to a wide range of scales in brain architecture such reconstruction requires imaging that is both high-resolution and high-throughput. Existing electron microscopy (EM) techniques possess required resolution in the lateral plane and either high-throughput or high depth resolution but not both. Here, we exploit recent advances in unsupervised learning and signal processing to obtain high depth-resolution EM images computationally without sacrificing throughput. First, we show that the brain tissue can be represented as a sparse linear combination of localized basis functions that are learned using high-resolution datasets. We then develop compressive sensing-inspired techniques that can reconstruct the brain tissue from very few (typically 5) tomographic views of each section. This enables tracing of neuronal processes and, hence, high throughput reconstruction of neural circuits on the level of individual synapses.*

## 1. Introduction

The complexity of human behavior is matched by the complexity of underlying neuronal circuits, reconstructing which is an important step towards understanding brain function [1]. Such reconstruction poses a formidable challenge because of the large numbers and the wide range of spatial scales involved. The human brain contains $10^{11}$ neurons, and each neuron receives and sends signals to nearly $10^4$ other neurons. The signals are transmitted along processes called neurites and across contacts called synapses. Circuit reconstruction requires identifying synapses and tracing neurites from synapses to neuron bodies.

The wide range of scales characteristic of brain architecture requires imaging with high resolution and high throughput. Because neurites belonging to different neurons are tightly packed and have similar composition, tracing them requires resolving neuron membranes. As neurites are often only a few tens of nanometers in diameter, resolving neuron membranes requires imaging with voxel size of about $10 \times 10 \times 10$ nm$^3$ [1]. At the same time, neurites are often many millimeters or centimeters long. Even a tiny fruit fly brain contains $2 \times 10^{13}$ such voxels thus requiring high-throughput imaging.

Because the required resolution is finer than the wavelength of light, neuronal circuit reconstruction must use electron microscopy (EM). The oldest EM technology with proven capability to reconstruct neuronal circuits is serial section Transmission EM (ssTEM) [2, 3]. While ssTEM achieves required resolution in the x-y plane, its depth resolution is limited by physical section thickness (about 50 nm) (Figure 1a). This limitation introduces two sources of errors that affect the ability to automatically trace neurites. First, while vertically oriented membranes are sharp and identifiable in ssTEM images, tilted membranes appear blurred because each ssTEM image is a projection of the section volume onto a plane, Figure 1a and b. Second, topologically distinct configurations of membranes may appear identical in a projected image [4]. These errors necessitate enhancement of the depth resolution of ssTEM.

Other existing EM technologies, differing in the way sections are cut and imaged, may offer high depth resolution or high throughput but not both. In particular, Scanning Transmission EM (STEM) and Scanning reflection EM (SEM) [1, 5] suffer from the low depth resolution like ssTEM. Alternatively, Focused Ion Beam (FIB) [6] or serial section electron tomography [7] possess high depth resolution but are limited in throughput. Tomography, in particular, is slow because it requires acquiring hundreds of different angle views of the same section.

In this paper, we propose a new EM technology that combines a high depth-resolution of electron tomography and FIB with high throughput of ssTEM, STEM, or SEM by relying on computational processing (Figure 2). The main idea is to image thin sections, either in transmission or reflection mode, but only in a few, say five, tilt views. Such limited tomography can achieve required resolution while sacrificing imaging speed only by a small number of tilt views. Previously, this idea was implemented using a man-made dictionary comprised of Gaussian-smoothed cuboids [4]. The length, width and thickness of these cuboids were chosen manually to represent the neuron membranes. To better capture the statistics of neuron membrane structures, in this paper, we learn the dictionary in an unsupervised fashion from experimental datasets acquired with low throughput but high resolution.

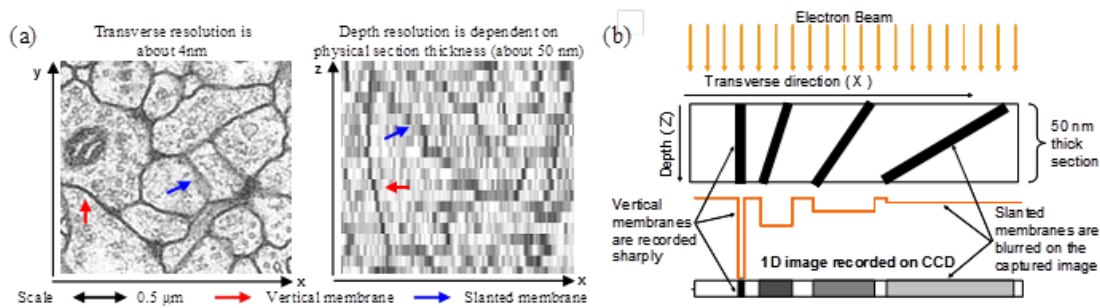

Figure 1: Limited depth resolution of conventional serial section imaging causes reconstruction errors a) Illustration of the x-y and z resolution of ssTEM for 50 nm thick sections. While x-y resolution is sufficient to follow neural processes running in z-direction, z-resolution, limited by physical section thickness, precludes tracing processes running in x-y plane. b) Schematic illustration of ssTEM imaging of membranes. Slanted membranes appear blurred on the captured image.

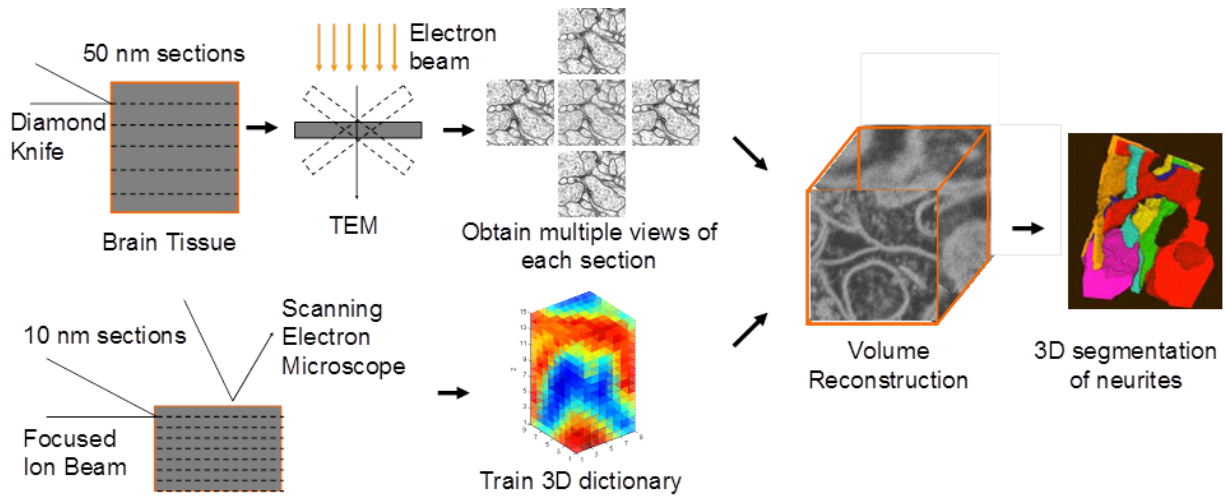

Figure 2: A block diagram of the proposed method to increase depth resolution for high throughput neural imaging. The brain tissue is first physically sliced into thin 50 nm sections using a diamond knife. Each section is then imaged several times with a different orientation of the section for each image. For this paper, we took five images: one normal view, and four views at 45° tilt of the sample in each direction about the x and y axes respectively. High resolution (10 nm isotropic) images for a smaller volume from the same brain region are collected using FIB to train an over-complete 3D dictionary. A computational reconstruction algorithm then uses the images for several sections together with priors about the representation of brain tissue over the learned dictionary to reconstruct the volume (to enhance visualization, grayscale contrast is reversed here and through the rest of the paper). 3D segmentation is applied to the volume in order to trace neural processes and identify synapses.

### 1.1. Contributions

This paper makes the following technical contributions:
• We address a major impediment towards automated high-throughput neural circuit reconstruction - low depth resolution of thin-section electron microscopy undermining the ability to trace neurons across sections.
• We show that the brain ultra-structure can be represented as a sparse linear combination of 3D dictionary atoms learned on experimentally acquired high-resolution datasets in an unsupervised fashion.

• We show that sparse reconstruction from very few tomographic views of each section can improve depth resolution several-fold both using simulated ground truth datasets and actual transmission EM images.

• We show that sparse reconstruction improves depth resolution even if only normal views (images acquired by the electron beam perpendicular to the brain section, Fig. 4) are available, making our method applicable to past and future datasets acquired on existing microscopes without modifying them for the acquisition of tilt views.

• We show that brain structure missing due to section folds or lost sections can be largely recovered from adjacent sections using 3D inpainting based on sparse representation over learned dictionary.

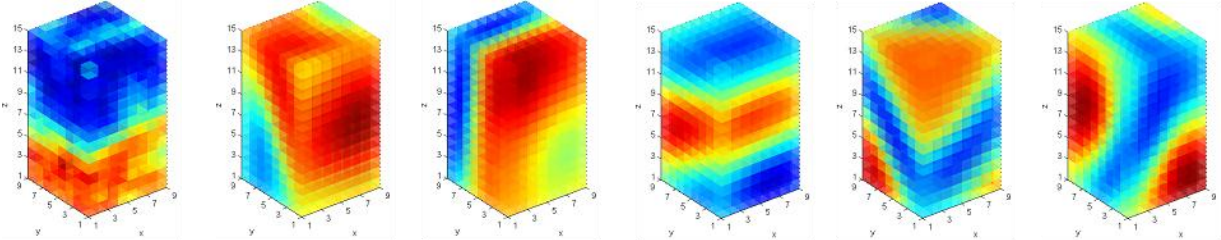

Figure 3: Representative 3D atoms from a dictionary trained on FIB data. For better visualization, jet colormap was used instead of grayscale.

## 2. Brain ultra-structure can be represented as a sparse linear combination over a learned over-complete dictionary

Analysis of natural images showed that they can be represented as a sparse linear combination of atoms of an over-complete dictionary [8]. This dictionary may be man-made, perhaps a Haar or DCT dictionary. Similarly, brain ultra-structure admits decomposition into a sparse linear combination of man-made atoms [4]. Most of the atoms model neuronal membrane patches as smoothed cuboids of various orientations and locations [4]. The advantage of such man-made dictionaries lies in their simple analytical formulations and fast implementations. However, typically, their ability to model complex naturalistic signals is handicapped by the over-simplification. This leads us to an alternative approach, where the dictionary is learned from data.

In a seminal paper [9], Olshausen and Field showed that an over-complete dictionary can be learned from natural image patches. Remarkably, the learned dictionary atoms look qualitatively similar to receptive fields of simple-cells in visual cortex. Their work demonstrates the potential of learning data structures from examples and has attracted much attention recently [10, 11]. The learned dictionaries can be adapted for specific classes of signals, and thus may better capture the statistical nature of the data. As demonstrated in various image processing applications [10, 11], sparse representation with learned dictionaries outperforms predefined dictionaries.

Here we adapt the idea of sparse over-complete representations [9-13] to the problem of brain reconstruction. Although [4] showed great improvement in depth resolution via using a man-made dictionary, learning a dictionary adapted for specific neural circuit volume may have the potential to yield better performance.

We attempt to represent volume patches of size $n = h \times h \times v$ voxels, vectorized as $X = \{x_i \in \mathbb{R}^n, i = 0, 1, 2...\}$, as linear combinations of very few ($<< n$) atoms of some over-complete dictionary $D \in \mathbb{R}^{n \times k}$ with $k > n$ atoms. The approximate representation can be written as

$$X \approx DA, \qquad (1)$$

where $A = \{a_i \in \mathbb{R}^k, i = 0, 1, 2...\}$ is an unknown coefficient matrix with only a few non-zero entries per column. The required dictionary which yields sparse representation of $X$ can be learned from solving a joint optimization problem with respect to the dictionary $D \in \mathbb{R}^{n \times k}$ and the sparse representation coefficients $A$:

$$\min_{D,A} \sum_i \frac{1}{2} \|x_i - Da_i\|_2^2 + \lambda \|a_i\|_1, \qquad (2)$$

where $\lambda$ is a regularization parameter controlling the trade-off between representation error and representation sparseness. The scaling ambiguity of $D$ is removed by constraining the atom length to norm one. Although such formulation is not jointly convex, it is convex with respect to each of the two variables $D$ and $A$ with the other fixed. Several algorithms have been proposed to perform dictionary learning by performing an iterative optimization alternatively over *D* and *A* until convergence [10, 12]. Neither dictionary atoms nor the weights of individual atoms in the representation are uniquely determined. However the goal of this work is not to find a unique decomposition but to produce an accurate high-resolution reconstruction. In sparse redundant representations, multiple combinations of coefficients and dictionary atoms can yield the same or very similar reconstruction [12]. Thus non-uniqueness of the decomposition is not an issue for reconstruction purposes.

Here we used publicly available SPAMS package [12] to train a dictionary applied later for recovering the depth resolution lost in the ssTEM data. In particular, we first collected high resolution data using FIB for a small volume of fly larva brain tissue with 10 nm isotropic resolution. Then we extract $9 \times 9 \times 15$ volume patches from the FIB volume data with overlaps of 5 voxels between adjacent patches. We choose the volume patch size as a compromise between the capability of representing typical neuron circuit structures and the complexity of computation. The patch size in z-direction is specially designed to be 15 voxels or 150 nm. With this setup, when the brain tissue is cut to 50 nm thick sections for ssTEM imaging, the projection result for each patch will contain the information from 3 consecutive sections. We will show later that the correlation between voxels from adjacent sections is important to filling in the missing voxels due to section folds (Figure 8). Empirically, we found that using large number of examples (one million volume patches) and of dictionary atoms improves reconstruction results. We compared reconstruction quality for complete, 1.5x, 2x and 3x over-complete dictionaries. The 2x over-complete dictionary yielded better quality compared to the smaller dictionaries, while the 3x over-complete dictionary was not much better than 2x. Thus, to balance performance and computational complexity, in our implementation, we set *D* to be twice over-complete with $k = 2n = 2430$ atoms and use $\lambda = 0.1$, which yield satisfactory reconstruction results as shown in the following sections.

A few representative atoms from the resulting dictionary are shown in Figure 3. Similarly to Gabor patches learned from 2D natural image patches[9], these atoms contain patches of membranes. Each training or test volume patch is sparsely represented by about 90 out of 2430 atoms with an average representation error $\|x_i - Da_i\|^2 / \|x_i\|^2 \approx 0.03$. Thus, the brain ultra-structure is a sparse linear super-position of dictionary atoms learned from examples.

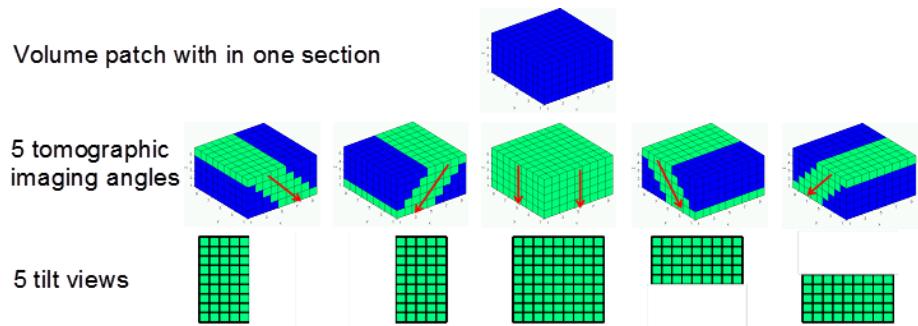

Figure 4: Schematic representation of the limited ssTEM tomography. Top row: 1/3 of the volume patch of $9 \times 9 \times 5$ voxels belonging to one section. Middle row: projections of the volume along 5 tomographic angles, only voxels (green) which are projected with voxels exclusively from the same patch are included. Projection angles are shown by red arrows. Bottom row: tilt views calculated as an average of voxels along the direction of projections.

## 3. Reconstructing brain ultra-structure from projections

In ssTEM, the desired $10 \times 10 \times 10$ nm$^3$ voxels are not imaged directly. The imaging results are projections of the 50nm thick section volume along different tomographic angles, e.g. the normal view image is the z-projection of the section, Figure 4. The goal is to recover the high resolution volume from very few tomographic views or even from a single normal view. Then we are facing the problem of solving an underdetermined system where the number of unknown voxels is greater than the number of measurement results. The inevitable redundancy among different tomographic views makes the system even more under determined. Although ill-conditioned, the system is still soluble with the prior knowledge that each volume patch can be sparsely represented over the learned dictionary.

Specifically, for a volume patch $x_i \in \mathbb{R}^n$, its projection measurements $y_i \in \mathbb{R}^m$ ($m < n$) can be written as

$$y_i \approx Px_i \approx PDa_i, \qquad (3)$$

where $P \in \mathbb{R}^{m \times n}$ ($m < n$) is a projection matrix approximating the ssTEM imaging process, and $a_i \in \mathbb{R}^k$ is the sparse representation coefficient vector. It is well known from the Compressive Sensing literature [14-17], if the matrix product $P \times D$ satisfies certain incoherent conditions, $a$ can be recovered near perfectly by solving the following equation

$$a_i = \arg\min_{a_i} \frac{1}{2} \|y_i - PDa_i\|_2^2 + \lambda \|a_i\|_1, \quad (4)$$

where $\lambda$ is a parameter used to balance the representation fidelity and the sparse prior promoted by the $l_1$ norm $\|a\|_1$. Putting back $a_i$ into $x_i' = Da_i$, the volume patch with high z-resolution is recovered. The brain tissue volume can be thought as a collection of overlapping volume patches $X = \{x_i \in \mathbb{R}^n, i = 0, 1, 2...\}$. In principle, one can use the tomographic views for the entire volume as a whole to solve for the coefficients associated with all patches simultaneously [4, 18]. In practice, due to the limit of computational capability, following [19, 20], we reconstruct each patch independently and restore each voxel of the original volume as a simple average of overlapping patches containing this voxel.

To apply the above mentioned method, we use the learned dictionary from Section 2 and construct the projection matrix $P$ based on the assumed tomographic imaging process shown in Figure 4. Since the projection procedure for each one of the three consecutive sections of the volume patch is identical, we only show 1/3 of the volume patch within one section. We assume each section contains five 10 nm thick voxel layers, and the ssTEM imaging measures the average voxel intensities along the direction of projection. Note, in the patch based method, each patch is treated independently, the projections which mix the boundary voxels (blue voxels in the middle row of Figure 4) with voxels from adjacent patches should be discarded from the tomographic results. From each 1/3 of the volume patch, we obtain 261 measurements (Figure 4 bottom row), which results in a projection matrix $P \in \mathbb{R}^{783 \times 1215}$ for the whole volume patch.

**Algorithm 1**: Super-resolution Reconstruction of Brain Structure in EM Using Sparse Representation over Learned Dictionary

**Input:** High resolution FIB data, low depth resolution ssTEM tilted views from 5 angles defined in Fig. 4, projection matrix $P$.

**Processing steps:**
**1:** Learning sparse representation dictionary $D$
Train a dictionary $D$ using $9 \times 9 \times 15$ volume patches $X$ from the high resolution FIB data with 5 voxel overlap in each direction. Solve the optimization problem

$$\min_{D,A} \sum_i \frac{1}{2} \|x_i - Da_i\|_2^2 + \lambda \|a_i\|_1.$$

**2:** Recovering high resolution volume from low resolution projections
Extract $9 \times 9$ patches of the ssTEM tilted views with 8 voxel overlaps in the x-y plane. Concatenating the patches from three consecutive sections and keeping only the green pixels as shown in Fig. 4 to obtain the projected view of $9 \times 9 \times 15$ patches. Find the sparse representation coefficients of each projected patch $y_i = Px_i$ by solving

$$\min_{a_i} \frac{1}{2} \|y_i - PDa_i\|_2^2 + \lambda \|a_i\|_1.$$

Generate the high resolution patch $x_i' = Da_i$, and restore each voxel of the high resolution volume by averaging the overlapping patches.
**3:** Smoothing the recovered volume
Extract $9 \times 9 \times 15$ patches of the recovered volume in step 2 with 8 voxel overlaps in each direction. Find the sparse representation for each patch by solving

$$\min_{a_i'} \frac{1}{2} \|x_i' - Da_i'\|_2^2 + \lambda \|a_i'\|_1.$$

Generate the smoothed patch $x_i'' = Da_i'$ and obtain the smoothed volume by averaging.

**Output:** Denoised high resolution volume.

Figure 5. The algorithm for super resolution reconstruction of brain structure using electron microscopy.

To minimize the potential artifacts at the patch boundaries when they are averaged to form the reconstructed volume, we densely sample the tomographic views to obtains a set of measurement results $Y = \{y_i \in \mathbb{R}^m, i = 0,1,2...\}$, from which overlapping patches with 1 voxel shift in the x-y plane and 5 voxel shifts in the z-direction are reconstructed. The minimum shifts in z-direction are determined by the fact that each ssTEM section is imaged separately and contains five 10 nm thick voxel layers. Because of this coarse sampling in the z-direction, some discontinuities are observed at the interface between every five layers of the reconstructed volume. To correct these artifacts, we resample the reconstructed volume densely and obtain a new set of patches $X'$ with 1 voxel shift in every direction. The patches are decomposed to sparse representations via solving the following equations,

$$\forall i \quad \min_{a_i'} \frac{1}{2}\|x_i' - Da_i'\|_2^2 + \lambda \|a_i'\|_1. \quad (5)$$

Then the smoothed patches are calculated as $x_i'' = Da_i'$, which are put back to form the final reconstructed volume. This step is essentially similar to the application of image denoising using sparse and redundant representations [11]. The entire process is summarized in Figure 5, where the optimization problems are solved using SPAMS [12].

## 4. Experiments on simulated ssTEM from FIB Data

In this section, we provide a proof of principle by applying the patch-based reconstruction method described in the previous section to the simulated ssTEM obtained from FIB data of fly larva brains, which are different from the data used to train the dictionary. We simulated ssTEM imaging of 50 nm thick sections by using high resolution FIB data and projecting every five consecutive voxel layers onto a single section. The existence of ground FIB truth for the simulated reconstruction allows us to evaluate its quality.

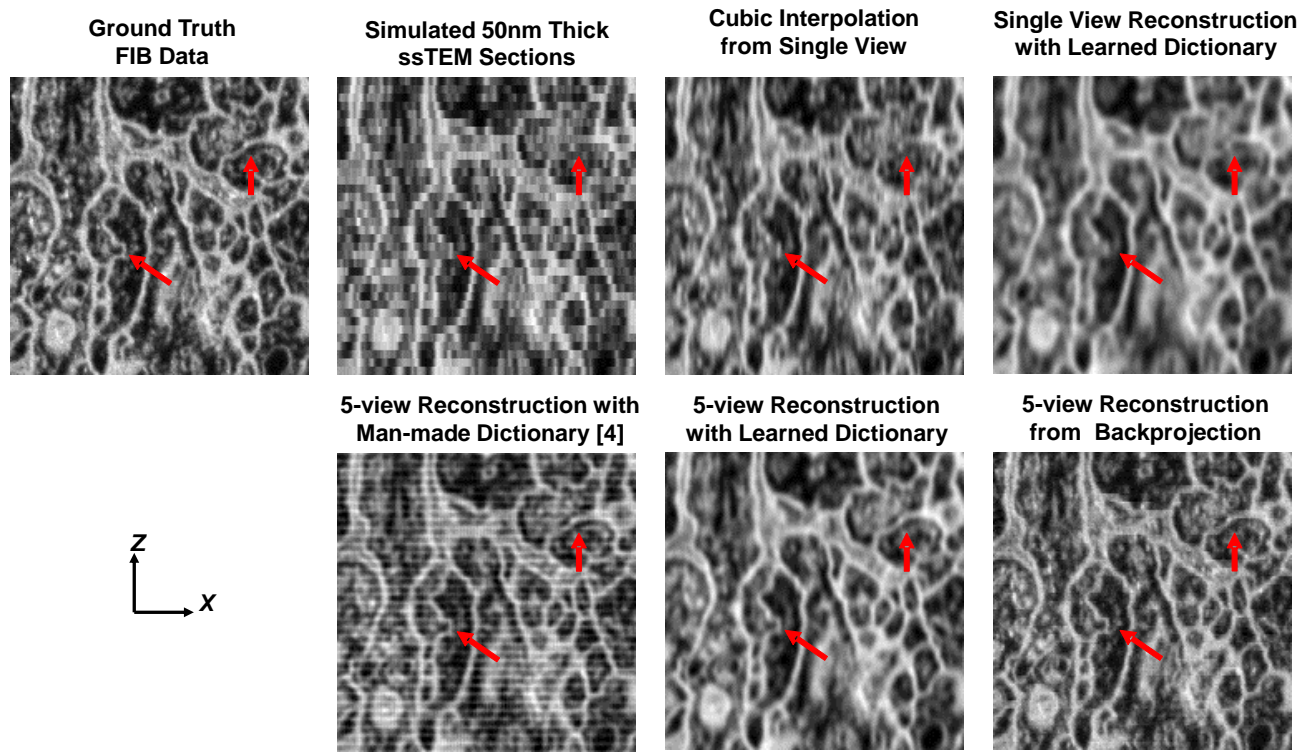

Figure 6: Representative examples of x-z reslice of the same region. Fine structures recovered by increased depth resolution are indicated by red arrows.

Representative examples of the x-z re-slices of the reconstructed volume using proposed method from 5 tilt views or a single central view are shown in Figure 6. The results clearly show improvement in z-resolution relative to the simulated ssTEM - see regions highlighted by red arrows. Interestingly, even a single-view reconstruction improves resolution substantially although not as much as with 5 tilt views. We also included other reconstruction methods in the comparison. Naïve cubic interpolation performs the worst. Reconstruction using a man-made dictionary [4] induces artifacts at the boundaries between adjacent sections and is sensitive to registration error (Figure 9). Backprojection reconstruction performs well in this experiment, but suffers from both noise (Figure 7) and registration imperfection (Figure 9).

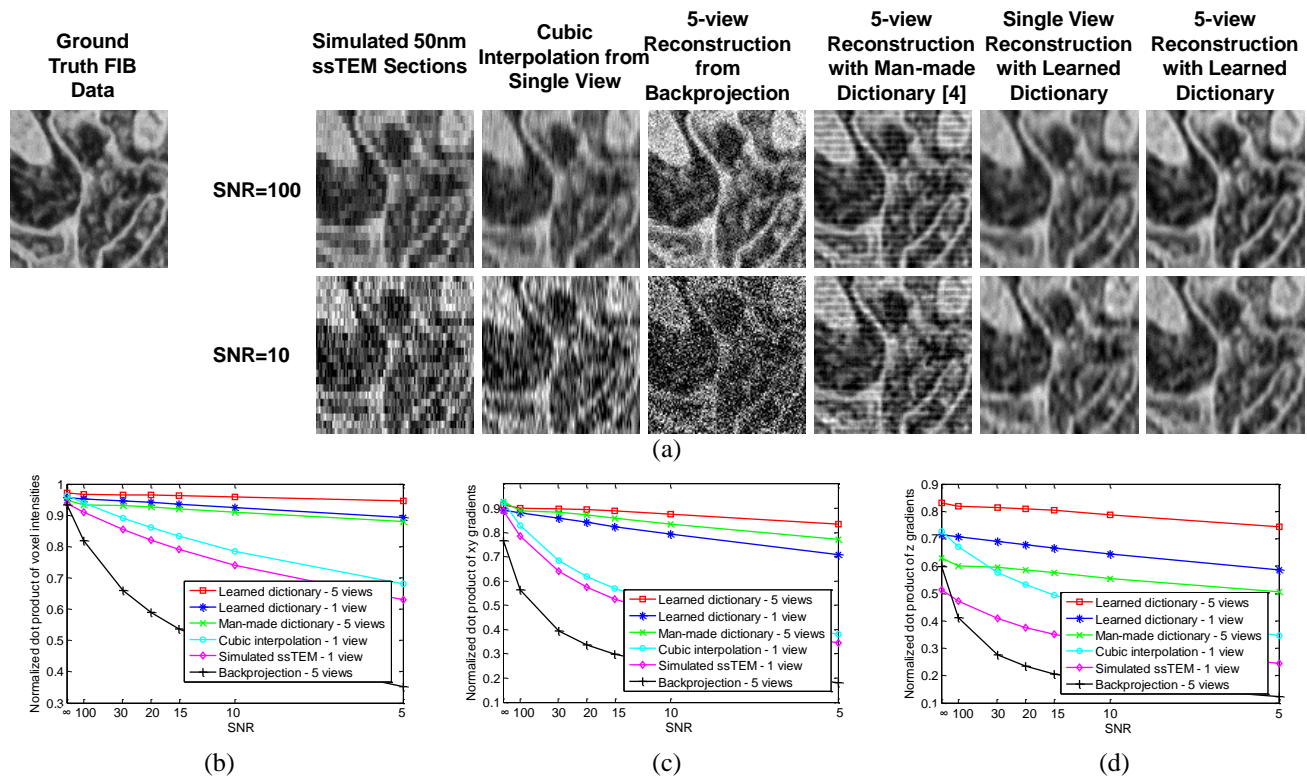

Figure 7: Performance of reconstruction with varying SNR. a) Representative examples of x-z re-slice of the same region. b) Normalized dot product between the true and reconstructed volume. c)-d) Normalized dot product between the true and reconstructed volume gradients in the x-y plane and z-direction repectively. Note that the thickness of the physical section affects the depth gradients $G_z$ much more significantly than the transverse gradients $G_{xy}$. $G_z$ is important to track membranes across sections. Our reconstruction captures $G_z$ and is robust to noise.

The requirement of high throughput limits the number of available electrons per pixel in ssTEM, and gives rise to electron shot noise. To demonstrate the robustness of our method to noise, we corrupted the tomographic views of the thick sections with Gaussian noise, and performed our reconstruction at various signal to noise ratio (SNR). We optimize the values of regularization parameter $\lambda$ for each SNR on a training dataset by exhaustively sampling the range of $\lambda$ and finding the value that gives quantitatively best results. Typical x-z re-slices of the reconstructed volume from 5 tilted views are shown in Figure 6a. The figure demonstrates that the reconstruction quality degrades slowly as the noise increases for our method and the approach using man-made dictionary [4]. As a comparison, the reconstruction obtained from 3D cubic interpolation or backprojection reconstruction degrades severely with noise. We also quantitatively evaluate the reconstruction using normalized dot product between the reconstructed volume and the ground truth volume, Figure 7b. The results clearly show that our method is capable of performing denoising and super-resolution reconstruction simultaneously and the learned dictionary outperforms the man-made dictionary [4].

Since the ultimate goal is to trace neurites and the ability to perform segmentation and tracing is dependent on the quality and smoothness of the gradients $G_x$, $G_y$ and $G_z$ within the volume, we further evaluate the quality of reconstruction by calculating the normalized dot product between the true gradients and the reconstructed gradients, Figure 7c, d. Compared to the ssTEM thick sections with low depth resolution or other existing reconstruction methods, our reconstruction significantly improves the z-gradient which enhances our ability to segment the volume and trace neural processes across sections.

Finally, we evaluate the reconstruction results by comparing segmentation of reconstructed volumes, Figure 8.

Specifically, we apply a 3D segmentation algorithm based on watershed [21] to the reconstructed volumes and to the original FIB data and compare the results with the ground truth boundary map generated from proofreading the 3D segmentation of the original FIB data. In addition, we compare ground truth with the 2D segmentation of the simulated ssTEM data because 3D segmentation is impossible in that case due to low depth resolution. Although the segmentation of the original FIB data performed the best, the 3D segmentation on our reconstructed volume from 5 simulated tilts performed better than 2D segmentation from simulated ssTEM. Our method performed the best among all reconstruction methods tested. Interestingly, even the 3D segmentation on the reconstructed volume from a single normal view shows significant improvement over the conventional 2D segmentation on the simulated ssTEM normal views.

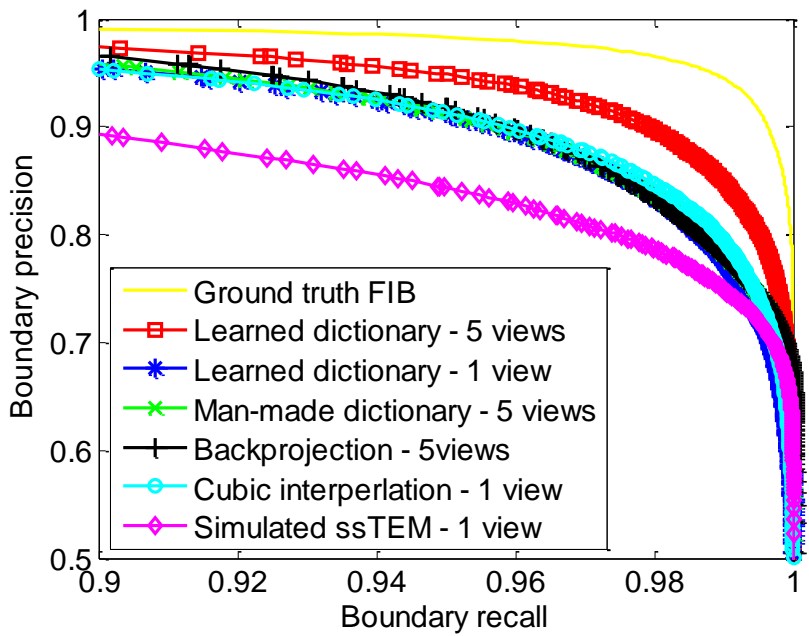

Figure 8: Results of 3D segmentation of the reconstructed volumes. Volume reconstructed from 5 simulated tilts or even a single normal view segments better than the 2D segmentation on the simulated ssTEM normal views.

## 5. Experiments on actual ssTEM data

To demonstrate that our method works on real data, we imaged fifteen consecutive 50 nm thick sections of fly larva brain with five tilts (normal view, 45° and −45° about the x-axis and y-axis) using a Tecnai Spirit electron microscope at 4 nm x-y pixel size. Under our microscope, the acquisition time of a single view of a 9 by 9 mosaic of 4000 ×4000 images is about 4 minutes, and it increases approximately linearly with the number of views. We also obtained high depth resolution training data using FIB, which is about 20 times slower than the ssTEM and has a much smaller field of view. Since even a tiny fly larval brain contains a few thousand sections, in order to complete imaging in reasonable amount of time, we have to use only a few angles for ssTEM and limit FIB to a much smaller volume used for training the dictionary. Although the present choices of the number of views and the tilt angles are not fully optimized, they satisfy our demands for both throughput and resolution. In the future, to speed up both imaging and computer processing, we may optimize these parameters by imaging each section from three orthogonal directions whose angles with respect to the norm of the section are identical.

The tilt images of each section were registered to the normal view by finding an affine transform that maximizes cross-correlation [22]. Normal views of each section were registered to each other using the same algorithm. The aligned images were down-sampled to 10 nm x-y pixel size for the reconstruction.

Representative x-z re-slices of the original thin sections imaged by ssTEM and the corresponding reconstructions are shown in Figure 9. Notice the significantly higher z-resolution in the reconstructions compared to the original ssTEM data, even for the case of reconstruction from single central view. Although not as good as the reconstruction with multiple tomographic views, single view reconstruction could be potentially valuable for the existing huge amount of ssTEM data collected only from central view and speed up the analysis of neural circuit volume which was hindered by the low z-resolution. Compared to simulations on FIB data, the remaining mis-alignment of real ssTEM sections degrade the reconstruction quality.

We also show in Figure 9 the volume slices reconstructed by using the method described in [4] with predefined dictionaries. The reconstructed data are further smoothed in z-direction. Compared to our method with learned dictionaries, this method is less robust against the inevitable misalignments between sections, which result in the artifacts at the section interfaces. Similarly, backprojection reconstruction also suffers from registration error.

We segmented both the original 50 nm thick ssTEM data and the reconstructed isotropic volume with a 10 nm depth resolution, then had a human expert proofread both segmentations. The proofreader reported that some areas could not be resolved in the ssTEM data but were trivial to resolve using orthogonal views in the isotropic data (Figure 10). Therefore, reconstruction of a complete connectivity will require technology to produce this kind of data with high isotropic resolution.

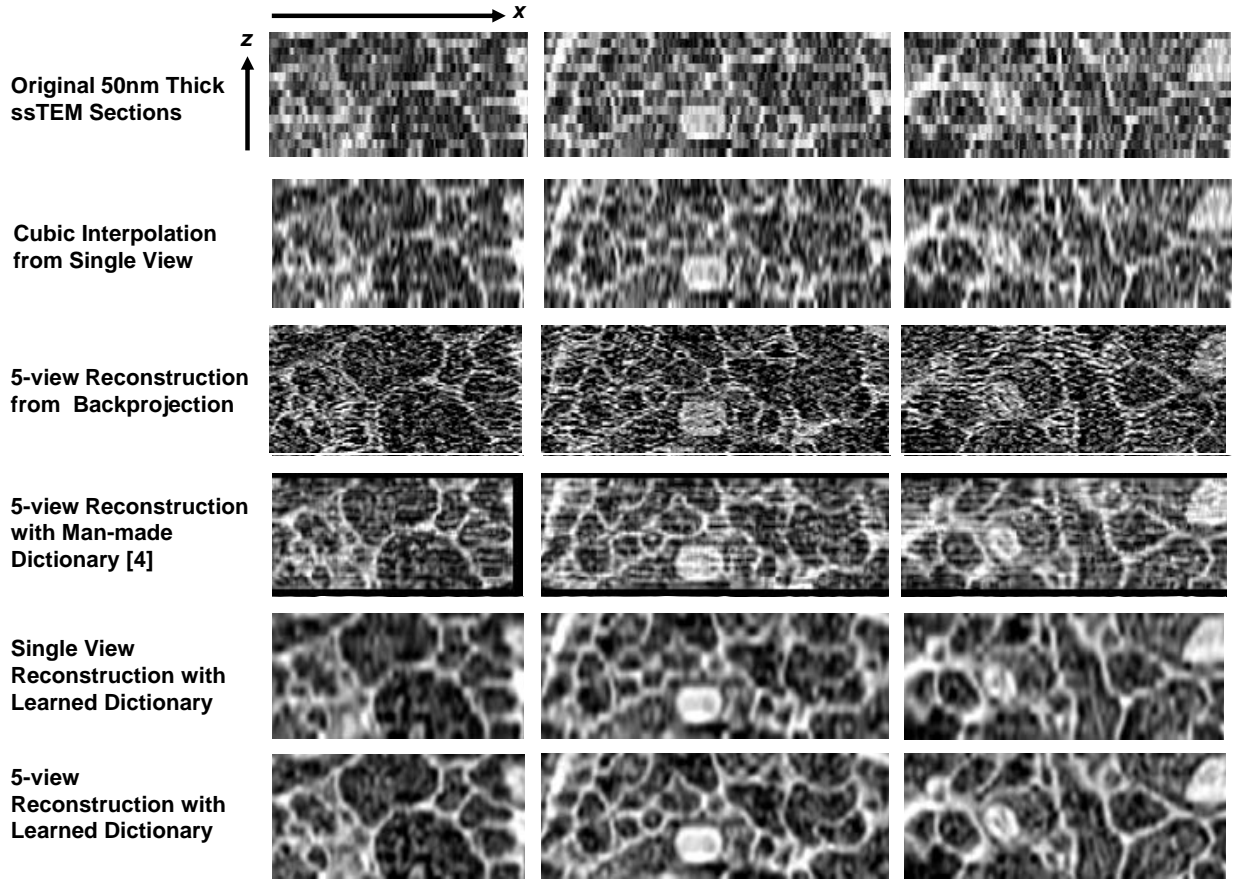

Figure 9: Experiments on actual data acquired via ssTEM. The top row shows x-z re-slices of an original ssTEM stack obtained at 50 nm section thickness. Notice the significantly higher depth resolution in the reconstructions with learned dictionary (fifth and sixth rows). The reconstruction with man-made dictionary [4] (fourth row) and the backprojection reconstruction (third row) are more sensitive than the proposed method to registration errors in actual data.

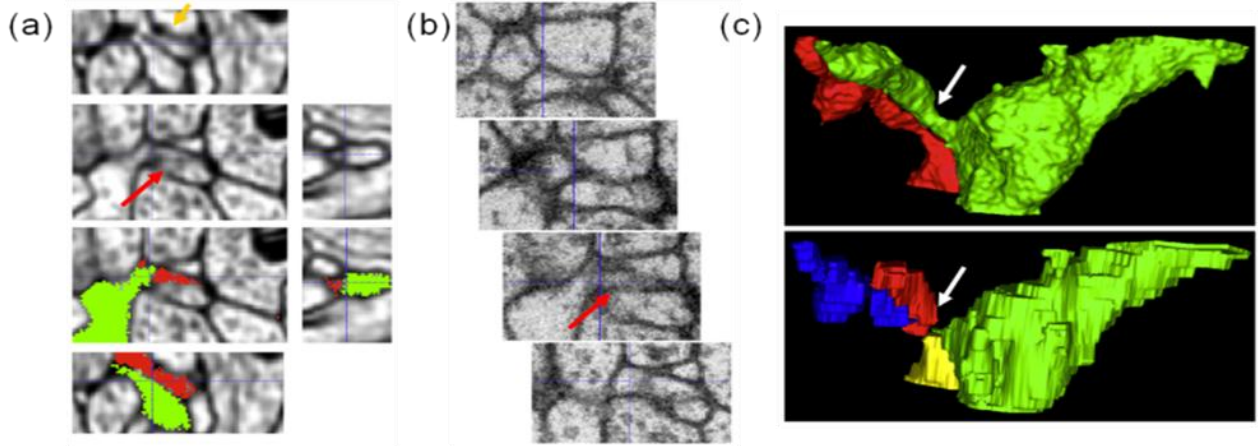

Figure 10: a) Difficult to reconstruct region in the isotropic volume, with orthogonal views and correct segmentation. Notice that the blurred area marked by the red arrowhead can be resolved in the orthogonal (x-z) view (orange arrowhead). b) Corresponding slices in the ssTEM data (slices +2, +1, 0 and -1, respectively), showing that the small processes seen in a) cannot be followed (red arrowhead marks the corresponding region). c) 3D visualization of proofread segmentations on the reconstructed volume (top) and ssTEM data (bottom). Notice that two bodies have been erroneously split because proofreaders could not follow the small processes in the ssTEM data.

## 6. Inpainting of folds

One of the artifacts of physically cutting the tissue into ultra thin sections is the presence of folds either in the section or in the support film (Figure 11b). Folds occlude neuronal structure complicating tracing of processes. We propose recovering missing voxels from the information available in adjacent sections. This is possible because folds appear independently on each section and are unlikely to lie directly on top of each other in consecutive sections. The same algorithm may be used to recover information from an isolated lost section.

We use a sparse representation over an over-complete 3D dictionary learned on FIB, in a process similar to filling in missing pixels of 2D images [10]. Specifically, in Equation (4), we ignore pixel values in the regions of images corresponding to folds. Although this results in fewer equations than in the fold-free case, we can still solve for the sparse coefficients, $A$, and use Equation (1) to recover the missing voxels as shown in Figure 11. The fold pixels were identified automatically. Since folds appeared much darker in the raw positive images, we thresholded pixel intensities at 4σ away from the mean image intensity. Next, we perform a dilation/erosion on the black pixels to get rid of tiny bridges across the folds. Then we search for connected components on the non-fold pixels, to rule out holes caused by specks of dust, etc.

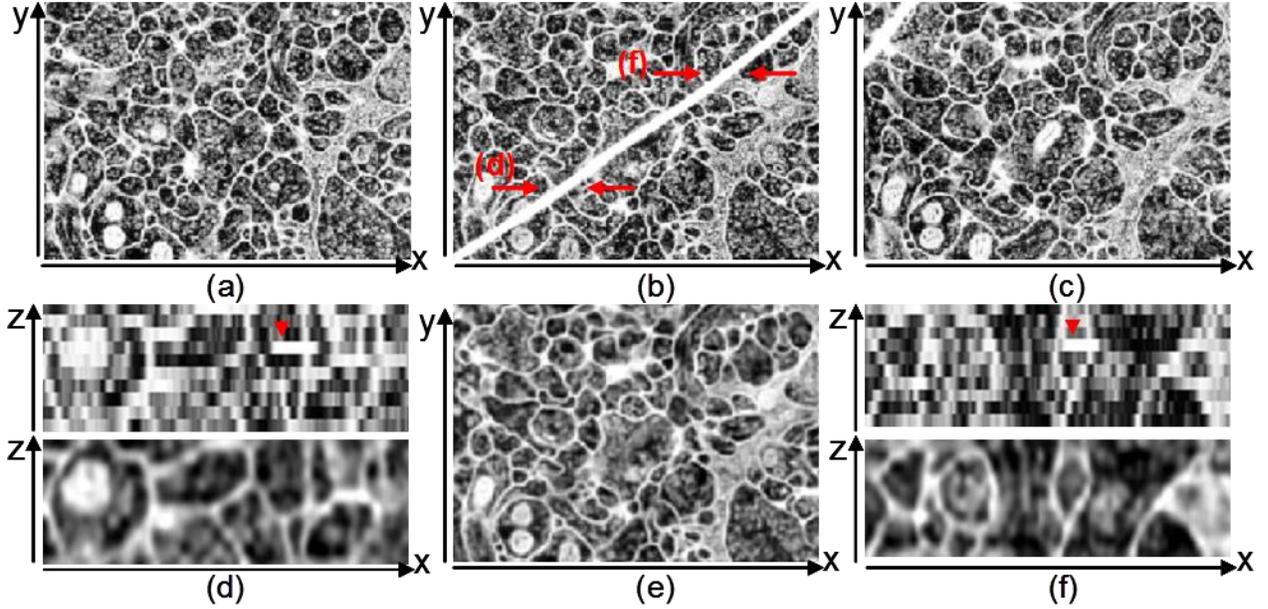

Figure 11: Filling in missing structure by 3D inpainting of folds. a)- c) normal ssTEM views of 3 consecutive 50 nm thick sections, among which the middle section (b) contains a fold. e) Average of the five 10 nm thick layers reconstructed for the middle section b). d) and f) Zoomed-in views of the x-z re-slices at the positions marked by arrow pairs in (b) and corresponding reconstructions, where fold locations are indicated by arrow heads.

## 7. Discussion

In this paper, we have proposed a method for increasing the depth resolution of ssTEM and recovering missing ultra-structures due to tissue folds using sparse representations over learned dictionaries. Given the sparse prior on the brain ultra-structure, we can learn a dictionary best suited for a particular brain tissue. Our reconstruction method requires only very few (typically 5) ssTEM images per section and therefore is only about a few times slower than traditional ssTEM while providing much higher depth resolution required for tracing neural circuits and identifying synapses. Other techniques which can provide similar depth resolution (such as serial section transmission electron tomography and FIB) are about tens of times slower which is a huge handicap since the goal is to image and reconstruct about $10^{13}$ voxels.

Our reconstruction approach extends previous work on 2D images [19, 20] to 3D. However, unsupervised learning is not the only way to attack the problem. Previously, supervised learning on convolutional networks was used for low-level vision tasks such as de-noising [23] and segmentation [24], albeit not for the problem presented here. Achieving good results with such supervised methods requires significant computational resources both because of the size of training datasets and because of the slow convergence in training. Interestingly, supervised learning can be accelerated by pre-training the network with the results of unsupervised learning [25, 26]. Thus, it will be interesting to see in the future what results a combination of unsupervised and supervised techniques would produce.

Our reconstruction approach exploits only information present locally in a patch. As demonstrated in [27, 28], by exploiting information available globally in an entire image, one may further improve reconstruction performance.

On the negative side, in contrast to FIB, we need to perform very accurate alignment and registration of the acquired images before they can be processed. Another possible source of error is the imperfect knowledge of the projection matrix *P*, which generates tomographic views by averaging the 3D volume of brain tissue in the direction of imaging. Instead of explicitly constructing *P*, one may use concatenated high and low resolution volume patches as training examples to learn simultaneously the coupled dictionaries [19, 20]. However, this would require re-imaging ssTEM sections with FIB, which has yet to be accomplished.

Although the computational complexity of the proposed method is high due to the large sizes of 3D patches and the dictionary [12], the problem is easily parallelizable. On a dual Quad Core X5550 workstation used for the simulations in the paper, the reconstruction time for a $200 \times 200 \times 200$ volume is about a day. Biologically interesting large volumes need to be divided into small blocks and reconstructed in parallel using a cluster. The training time for a $1215 \times 2430$ dictionary using 1.5 million examples is about two weeks. Although the best results are obtained by re-learning the dictionary for each brain region the learning can be sped up by initializing the learning algorithm using a previously learned dictionary.

In summary, we believe that the proposed method allows one to achieve the high-resolution and high-throughput imaging necessary for the reconstruction of brain circuits.